%% file: main.tex
\documentclass{article}

\usepackage[preprint]{neurips_2026}

\usepackage[utf8]{inputenc}
\usepackage[T1]{fontenc}
\usepackage{amsmath,amsfonts,amssymb}
\usepackage{booktabs}
\usepackage{caption}
\usepackage{colortbl}
\usepackage{enumitem}
\usepackage{float}
\usepackage{graphicx}
\usepackage[hidelinks]{hyperref}
\usepackage{microtype}
\usepackage{multirow}
\usepackage{nicefrac}
\usepackage{tikz}
\usetikzlibrary{arrows.meta,backgrounds,calc,fit,positioning}
\usepackage{url}
\usepackage{xcolor}
\usepackage{xspace}

\input{preamble}

\title{\method: Transferring Representations from Video Generation Models for Dynamic Spatial Reasoning}

\author{%
  Ke Yang$^{1}$ \quad
  Zhenyu Zhang$^{1}$\quad
  Jun Li$^{1}$
  \\[0.5em]
  $^{1}$Nanjing University \quad
}

\begin{document}

\maketitle

\input{Sec/abstract}
\input{Sec/intro}
\input{Sec/related}
\input{Sec/method}
\input{Sec/exp}
\input{Sec/conclusion}

{
  \small
  \bibliographystyle{ieeenat_fullname}
  \bibliography{main}
}

\end{document}

%% file: preamble.tex
\newcommand{\method}{\textit{GenDSR}\xspace}

%% file: Sec/abstract.tex
\begin{abstract}
Dynamic spatial reasoning is essential for connecting visual intelligence to
the physical world, yet remains challenging for vision-language models (VLMs),
which must preserve scene state while tracking how spatial relations evolve
over time.  Meanwhile, video generation models (VGMs), trained to synthesize temporally
coherent world dynamics, offer a rich source of
spatiotemporal representations.  We present \textbf{\method}, a
framework that transfers intermediate representations from VGM
into VLM for dynamic spatial reasoning.  Rather than treating a VGM
feature as a monolithic prior, \method constructs two complementary branches: a raw branch
preserving holistic generative context and a delta branch emphasizing temporal variation.
Task-conditioned tokenwise gates then fuse both signals and inject the result into the
native visual representation.  On DSR-Bench, \method achieves
\textbf{67.2\%}, surpassing the previous best result and establishing
a new state of the art.
\end{abstract}

%% file: Sec/intro.tex
\section{Introduction}
\label{sec:intro}
The physical environments in which intelligent systems operate are
constantly changing: objects move, viewpoints shift, and spatial relationships
evolve over time. Reasoning in such environments requires more than recognizing static
scene states—it requires tracking and understanding their changes across time.
Dynamic spatial reasoning, which relates observed motion to changes in position,
orientation, distance, and inter-object relations, is therefore fundamental to
intelligence in the physical world. Tasks such as navigation, manipulation, and
autonomous decision-making depend not only on understanding the
current spatial configuration, but also on anticipating how it
changes~\citep{zhou2026learning,huang2026thinking}.

Yet, despite substantial advances in vision-language models and video understanding,
dynamic spatial reasoning remains far from reliable: recent benchmarks reveal
persistent failures in integrating motion cues, disentangling camera and object
motion, and maintaining temporally coherent spatial estimates~\citep{liao2026spamembenchmarkingdynamicspatial,
zhou2025vlm4d,zhang2025dsi}.

A prominent line of recent work
enhances VLMs with spatial priors from pretrained 3D foundation models~\citep{zhou2026learning,zhang2026make}, such as
VGGT~\citep{wang2025vggt} and $\pi^3$~\citep{wang2026pi}. While their latent features encode
rich geometric structure, they are primarily learned for scene reconstruction
rather than modeling temporal evolution, and rely on geometry-centric training
data that are costly and difficult to collect at scale.

In contrast, video generation models~\citep{wan2025wan,yang2025cogvideox} provide a particularly suitable
source of priors for dynamic spatial reasoning. Specifically, (1)
\textbf{Spatial consistency.} To synthesize coherent videos under object and
camera motion, these models must preserve scene structure and object
identity across continuously changing viewpoints. Prior studies further suggest that
their internal representations capture meaningful spatial cues, including camera pose
and 3D structure, indicating spatial awareness beyond appearance-level
semantics~\citep{huang2026much,shen2026pretraining}.
(2) \textbf{Temporal dynamics.} Video generation explicitly requires modeling how
visual content evolves over time, including object motion, camera movement,
and interactions among entities. This objective naturally encourages representations sensitive
to temporal changes that are essential for reasoning about evolving
spatial relationships~\citep{ho2022video}.
(3) \textbf{Scalable learning.} Unlike geometry-centric models that
depend on specialized multi-view data and geometric supervision, video
generation models can learn directly from abundant real-world video
collections.
This provides a scalable source of both spatial and temporal
knowledge without requiring costly 3D annotations or reconstruction labels.
This naturally raises a question: can the representations learned by video
generation models serve as dynamics-aware priors for spatial reasoning?

To this end, we propose \method, a lightweight framework
that transfers generative representations into VLMs for dynamic spatial
reasoning. Our key insight is to expose two complementary views of the generative representation: its holistic spatiotemporal context and its explicit local variations.
Accordingly,
\method constructs generative representations with two complementary branches. (1)
\textbf{Raw branch.} We directly align the intermediate features of
the video generation model with the VLM visual space,
preserving rich spatial and spatiotemporal information about the scene.
(2) \textbf{Delta branch.} We explicitly capture temporal variations by
computing feature differences across neighboring observations, emphasizing changes induced
by object motion, viewpoint shifts, and evolving spatial relationships.
The two branches are jointly integrated through lightweight residual
fusion, complementing the semantic representations of the original VLM
with information about both spatial states and their temporal evolution.

On DSR-Bench~\citep{zhou2026learning}, \method improves the Qwen3-VL-8B
SFT baseline from 62.0\% to 67.2\%,
while the 4B variant improves from 58.8\% to 62.7\%. Raw and delta representations
independently improve the accuracy,
and their combination yields the strongest overall performance.
These results validate video generation models as a
promising source of dynamic spatial priors for vision-language reasoning.

Our contributions are summarized as follows:
\begin{itemize}
    \item We explore video generation models as a source
    of spatial and temporal priors for dynamic spatial reasoning,
    offering an alternative to geometry-centric representations.
    \item We propose \method, which constructs two complementary representations
from generative features: a raw representation of spatial states and a delta
representation emphasizing temporal variations, and efficiently integrates both into VLMs.
    \item Experiments demonstrate consistent improvements in dynamic spatial
reasoning and reveal the complementary roles of generative spatial context
and temporal variations.
\end{itemize}

%% file: Sec/related.tex
\section{Related Work}
\label{sec:related}
\subsection{Dynamic Spatial Reasoning}

Spatial reasoning in vision-language models has progressed from understanding
static layouts to reasoning about spatial configurations that evolve over
time~\citep{yang2025thinking,zhou2026learning}. Unlike static settings, dynamic
spatial reasoning requires models to integrate evidence across frames,
distinguish object motion from viewpoint changes, and maintain temporally
coherent estimates of position, orientation, and inter-object relations. Recent
studies show that even strong VLMs remain unreliable under these coupled spatial
and temporal demands~\citep{zhou2025vlm4d,huang2026thinking}.

Existing methods improve dynamic spatial reasoning through task-specific
post-training, explicit geometry, or structured internal representations.
SpaceR~\citep{ouyang2025spacer} combines large-scale video spatial data with
spatially guided reinforcement learning. DSR Suite~\citep{zhou2026learning}
formalizes dynamic spatial reasoning from
monocular videos and introduces geometry-aware training data and a
Geometry Selection Module that retrieves question-relevant information from pretrained
3D reconstruction priors. Building on geometry-centric representations, GeoSR~\citep{zhang2026make}
encourages VLMs to actively exploit geometry tokens through visual-token
masking and adaptive geometric fusion.
DynTrace~\citep{gao2026dyntracetrackingdynamicobject} instead maintains geometry-informed object
trajectories and structured dynamic traces at inference. Related work has also
explored learning internal geometric or dynamic representations for reasoning.
3DThinker~\citep{chen2025think}
aligns latent reasoning states with representations from a pretrained
3D foundation model, while 4DThinker~\citep{chen20264dthinker} introduces dynamic
latent imagery and reinforcement learning for spatiotemporal reasoning.

\subsection{Video Generation Models for Spatial Reasoning}

Recent evidence suggests that video generation models learn spatial structure
beyond the appearance and motion statistics required for synthesis~\citep{huang2026much,shen2026pretraining}.
Moving from analysis to transfer,
VEGA-3D~\citep{wu2026vega} extracts intermediate video-diffusion features as
implicit 3D priors and adaptively fuses them with MLLM visual tokens for scene
understanding. DyVA~\citep{zhang2025worldmodelsbenefitvlms} similarly repurposes a
video diffusion model as a generative encoder, querying it with a single
denoising step and injecting the resulting latents as auxiliary visual
embeddings. GeoWorld-VLM~\citep{gu2026geoworldvlmgeometryworldmodels} transfers geometry from a
camera-conditioned world model through feature distillation.
Another line of work turns generative models into explicit visual imagination.
MindJourney~\citep{yang2025mindjourneytesttimescalingworld} couples a VLM with a controllable world
model at test time and iteratively synthesizes views along model-proposed camera
trajectories. \citet{zhang2026thinkingnovelviewssystematic} systematically
studies this generate-then-reason paradigm. World2VLM~\citep{zhang2026world2vlmdistillingworldmodel}
instead uses trajectory-conditioned world-model generations as training-time
supervision, distilling forward and inverse spatial reasoning into the VLM.
Different from prior approaches that transfer generative
representations as a unified auxiliary feature,
GenDSR focuses on dynamic spatial reasoning and
explicitly exposes two complementary views of the
VGM representation: its raw spatiotemporal state and
its local temporal variation.
This decomposition allows the VLM to access temporal
changes directly rather than relying on them to remain
implicitly encoded in the raw generative feature.

%% file: Sec/method.tex
\section{Method}
\label{sec:method}

\subsection{Overview}
\label{sec:method_overview}

\begin{figure}[t]
    \centering
    \includegraphics[width=\linewidth]{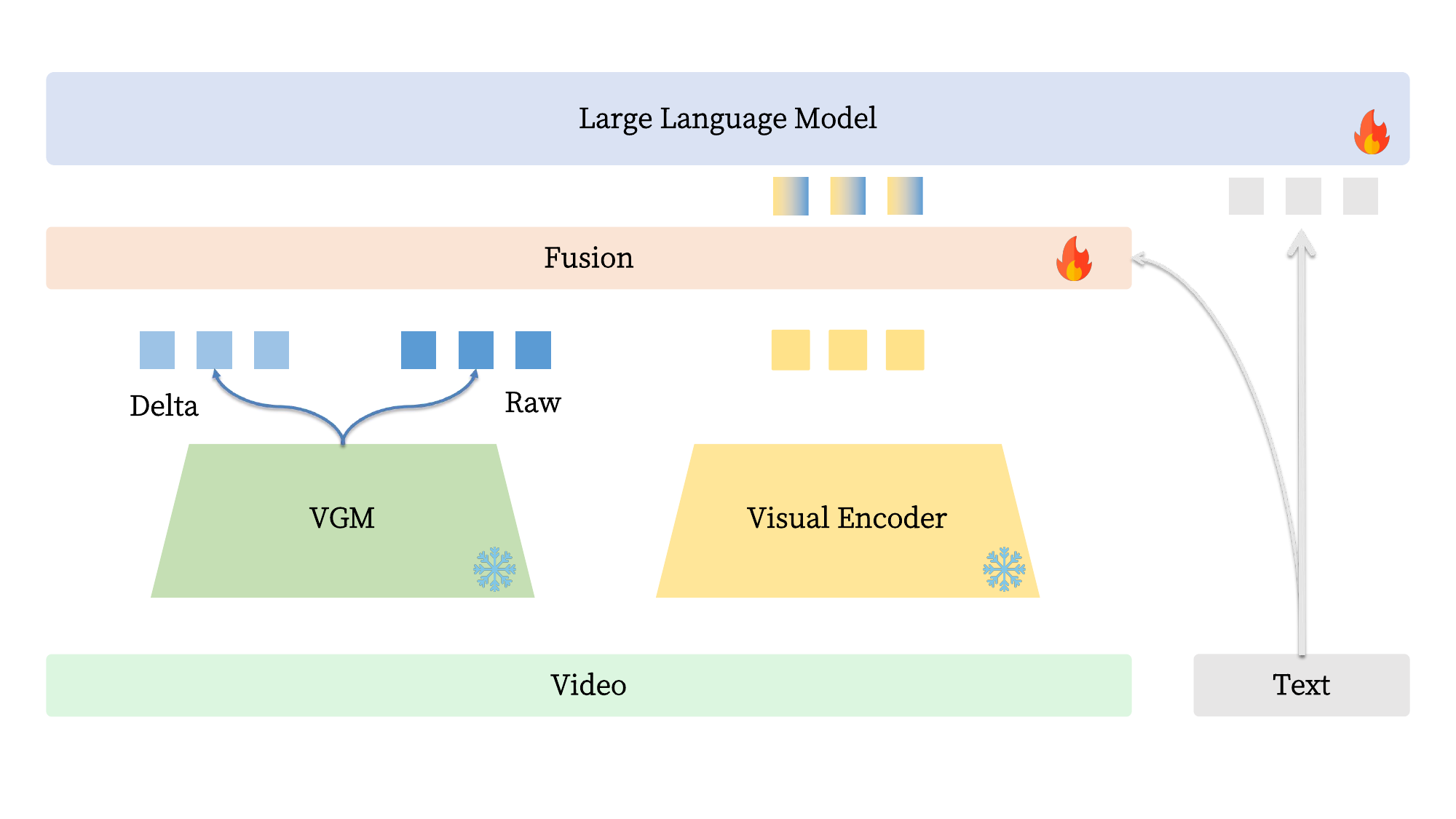}
    \caption{Overview of \method. The frozen native visual encoder and video
    generation model process the same input video. We construct raw and delta
    representations from the VGM features and integrate them with the native
    visual representation through task-conditioned fusion before language-model
    reasoning. Snowflake and flame icons denote frozen and trainable modules,
    respectively.}
    \label{fig:overview}
\end{figure}

Given a video $x$, a question $q$, and an answer
sequence $y$, a video-language model first encodes the video
into visual representations and predicts the answer autoregressively.
We denote the frozen visual encoder by $E_{\mathrm{vis}}$
and the native visual merger by $M_{\mathrm{vis}}$:
\begin{equation}
    H = E_{\mathrm{vis}}(x), \qquad
    S = M_{\mathrm{vis}}(H)
    \in \mathbb{R}^{N \times d},
    \label{eq:native_tokens}
\end{equation}
where $N$ denotes the number of visual tokens and
$d$ is the hidden dimension of the language model.  We associate these tokens
with a spatiotemporal grid $G=(T_v,H_v,W_v)$, such that
$N=T_vH_vW_v$.
Our goal is to enrich the native visual representation $S$
with complementary knowledge from a frozen video generation model
(VGM), while preserving the original visual pathway of the VLM.

Figure~\ref{fig:overview} provides an overview of \method.
We extract intermediate representations from the pretrained VGM and
construct two complementary representations from them:
a \textbf{raw branch} that preserves holistic generative context and
a \textbf{delta branch} that explicitly emphasizes temporal variations.
Both representations are aligned to the native VLM token grid
and projected into the language-model hidden space. A lightweight,
task-conditioned fusion module then determines their contributions at each
visual-token position and injects the resulting representation as a
residual immediately before the language model.

The following sections introduce the raw--delta representation transfer
in Sec.~\ref{sec:raw_delta}, the adaptive fusion mechanism in
Sec.~\ref{sec:gated_fusion}, and the training objective in
Sec.~\ref{sec:training_objective}.

\subsection{Raw--Delta VGM Representation Transfer}
\label{sec:raw_delta}

\paragraph{Generative feature extraction.}
Unlike a conventional visual encoder, a diffusion-based VGM operates through a
latent denoising process.
We first encode the observed video with the frozen
VGM VAE:
\begin{equation}
    z_0=E_{\mathrm{vae}}(x).
    \label{eq:clean_vgm_latent}
\end{equation}
For timestep $\tau$, we draw $\epsilon\sim\mathcal{N}(0,I)$ and apply the
native VGM scheduler's noising operator $\mathcal{N}_{\tau}$:
\begin{equation}
    z_{\tau}=\mathcal{N}_{\tau}(z_0,\epsilon),
    \label{eq:noisy_vgm_latent}
\end{equation}
where $\mathcal{N}_{\tau}$ follows the VGM's native noise schedule. We then perform one forward pass through the frozen denoising transformer
under the empty-prompt condition $c_{\emptyset}$.  Let
$\mathcal{D}_{\ell}(z_{\tau};\tau,c_{\emptyset})$ denote the intermediate
activation at block $\ell$, after patch embedding and propagation through the
preceding transformer blocks.  We extract
\begin{equation}
    Z=\mathcal{D}_{\ell}(z_{\tau};\tau,c_{\emptyset})
    \in\mathbb{R}^{T_g\times H_g\times W_g\times d_g},
    \label{eq:vgm_feature}
\end{equation}
where the first three dimensions form the native VGM feature grid and $d_g$ is
the VGM channel dimension.  The empty-prompt condition prevents an external
text prompt from contributing semantic information.  This extraction requires
only a single denoising-network forward at a fixed timestep, rather than an
iterative denoising trajectory.

The VGM grid $(T_g,H_g,W_g)$ generally differs from the VLM grid $G$.  We use
one shared, non-parametric alignment operator $\mathcal{A}_{G}$ for both
branches:
\begin{equation}
    \mathcal{A}_{G}:
    \mathbb{R}^{T_g\times H_g\times W_g\times d_g}
    \rightarrow \mathbb{R}^{N\times d_g}.
    \label{eq:grid_alignment}
\end{equation}
It interpolates the temporal axis to $T_v$, applies adaptive average pooling
to the target spatial grid $(H_v,W_v)$, and flattens the aligned grid into $N$
tokens.  The two branches use projectors with the same architecture but
independent parameters.  For $k\in\{r,d\}$, each projector is implemented as
\begin{equation}
    P_k(U)
    = W_{k,2}\,
      \operatorname{GELU}\!\left(
          W_{k,1}\operatorname{LN}_k(U)+b_{k,1}
      \right)+b_{k,2},
    \label{eq:branch_projector}
\end{equation}
where the first linear layer maps $d_g$ to $d$ dimensions and the second
operates in the $d$-dimensional language-model space.

\paragraph{Raw branch.}
The raw VGM representation retains the rich spatiotemporal context
learned through video generation. We first align its native
spatiotemporal grid with that of the VLM and then
project the aligned features into the language-model hidden space:
\begin{equation}
    R = P_r\!\left(\mathcal{A}_{G}(Z)\right)
    \in \mathbb{R}^{N \times d},
    \label{eq:raw_branch}
\end{equation}
where $P_r$ is a learnable feature projector. Rather than replacing
the native representation $S$, the resulting feature $R$ provides
complementary spatial and spatiotemporal context acquired from generative pretraining.

\paragraph{Delta branch.}
While the raw representation contains rich contextual information, temporal
changes may remain entangled with appearance and scene content.
We therefore explicitly expose temporal variations in the VGM
feature space by computing temporal differences between neighboring features:
\begin{equation}
    (\Delta Z)_1 = \mathbf{0}, \qquad
    (\Delta Z)_t = Z_t - Z_{t-1},
    \quad t = 2,\ldots,T_g,
    \label{eq:temporal_delta}
\end{equation}
where $T_g$ denotes the temporal resolution of the VGM
representation.  Prepending the zero feature preserves the original temporal
length, so $\Delta Z$ contains the same $T_g$ positions as $Z$.  Importantly,
temporal differencing is performed in the native VGM feature space before
spatiotemporal alignment, so temporal differences are defined directly on the
native VGM representation rather than on resampled features.

The resulting temporal differences are subsequently aligned and projected:
\begin{equation}
    D = P_d\!\left(\mathcal{A}_{G}(\Delta Z)\right)
    \in \mathbb{R}^{N \times d},
    \label{eq:delta_branch}
\end{equation}
where $P_d$ is a learned projector. The delta representation $D$ emphasizes
temporal variations associated with object motion, viewpoint changes, and
evolving spatial relationships, complementing the holistic context retained
by $R$.

\suppressfloats[t]
\input{Tabs/tab_dsr_bench}

\subsection{Task-Conditioned Tokenwise Fusion}
\label{sec:gated_fusion}

The usefulness of raw and delta information can vary across
questions and visual locations. For example, questions about static
spatial configurations may rely primarily on scene context, whereas questions
about motion or changing relationships may benefit more from temporal
variations. A fixed global mixture therefore cannot adapt to
the diverse requirements of dynamic spatial reasoning.

Let $\mathcal{I}_p$ index the valid prompt-side text tokens, excluding visual
tokens, target-response tokens, and padding, and let $e_j\in\mathbb{R}^{d}$ be the
corresponding input embedding from the VLM language embedding layer.  For our
multiple-choice prompts, $\mathcal{I}_p$ includes both the question and its
answer-option text.  We obtain the task representation by mean pooling:
\begin{equation}
    Q = \frac{1}{|\mathcal{I}_p|}
        \sum_{j\in\mathcal{I}_p} e_j
    \in \mathbb{R}^{d},
    \qquad
    \widetilde{Q}=\mathbf{1}_{N}Q^{\top}
    \in\mathbb{R}^{N\times d}.
    \label{eq:prompt_pooling}
\end{equation}
The input embeddings are detached only when constructing $Q$, which prevents the gate from introducing
an auxiliary gradient path through the prompt representation. We broadcast $Q$ over
the visual-token grid and condition the gate on the native visual tokens, both
VGM branches, and the resulting task context.  Specifically, we form
\begin{align}
    U ={}& [
        \operatorname{LN}_s(S),
        \operatorname{LN}_r(R),
        \operatorname{LN}_d(D),
        \operatorname{LN}_q(\widetilde{Q})
        ] \in \mathbb{R}^{N\times 4d},
        \\
    w_k ={}& h_k(U) = U W_k + b_k,
    \qquad k\in\{r,d\},
    \label{eq:gating_heads}
\end{align}
where $[\cdot]$ denotes feature concatenation and $h_r,h_d$ are independent
linear heads from $\mathbb{R}^{4d}$ to $\mathbb{R}$.  The unconstrained
coefficients allow each branch to be independently amplified, suppressed, or
sign-adjusted.  We initialize $W_r$ and $W_d$ to zero and their biases to
$b_r=1.0$ and $b_d=0.1$, respectively.

The fused generative residual is then computed as
\begin{align}
    F &= w_r \odot R + w_d \odot D,
    \label{eq:fusion_residual}\\
    S' &= S + F,
    \label{eq:residual_injection}
\end{align}
where $\odot$ denotes tokenwise scaling with broadcasting over the
hidden dimension. Only visual-token representations are modified; text-token embeddings
remain unchanged. The residual is injected after the native
visual merger and immediately before the language model, allowing
the generative features to complement rather than replace the native
VLM representation.

\subsection{Training Objective}
\label{sec:training_objective}

The augmented visual tokens $S'$ together with the question
$q$ condition the standard autoregressive supervised fine-tuning objective:
\begin{equation}
    \mathcal{L}_{\mathrm{SFT}}
    =
    -\sum_{i=1}^{L}
    \log p_{\Theta}
    \left(
        y_i
        \mid
        y_{<i}, S', q
    \right),
    \label{eq:sft_objective}
\end{equation}
where $L$ denotes the length of the target answer
and $\Theta$ contains all trainable parameters.  Both the VGM, including its
VAE and denoising transformer, and the native VLM visual encoder remain
frozen.  We optimize the language model, the native visual merger, the two
branch projectors, and the fusion module.

%% file: Tabs/tab_dsr_bench.tex
\newcommand{\up}[1]{\textsuperscript{\scriptsize\textcolor{gainclr}{$\uparrow$#1}}}

\begin{table*}[t]
\centering
\small
\setlength{\tabcolsep}{3.6pt}
\renewcommand{\arraystretch}{1.05}
\caption{\textbf{Dynamic spatial reasoning on DSR-Bench.}
Accuracy (\%) is reported for 13 subtasks and overall accuracy.}
\vspace{-0.3em}
\label{tab:dsrbench}
\resizebox{\textwidth}{!}{%
\begin{tabular}{l ccccccccccccc c}
\toprule
\textbf{Models}
& \textbf{A.Dis} & \textbf{A.Dir} & \textbf{A.Ori} & \textbf{A.Spd}
& \textbf{A.SpdC} & \textbf{A.DirP}
& \textbf{R.Dis} & \textbf{R.Dir} & \textbf{R.Ori} & \textbf{R.Spd}
& \textbf{R.SpdC} & \textbf{R.DirP}
& \textbf{N-Temp}
& \textbf{Avg. $\uparrow$} \\
\midrule
\multicolumn{15}{c}{\textbf{\textit{Proprietary Models}}} \\[1pt]
GPT-5
  & 21.1 & 41.5 & 48.7 & 34.5 & 33.3 & 34.7
  & 17.2 & 44.3 & 41.9 & 21.2 & 25.0 & 30.9
  & 26.7 & 30.8 \\
Gemini-2.5-Pro
  & 20.0 & 44.6 & 53.6 & 27.3 & 38.7 & 30.5
  & 23.2 & 32.9 & 43.2 & 17.1 & 28.5 & 27.9
  & 34.3 & 31.7 \\
\midrule
\multicolumn{15}{c}{\textbf{\textit{Spatial Reasoning Models}}} \\[1pt]
VLM-3R
  & 28.2 & 27.6 & 31.7 & 42.8 & 38.7 & 33.0
  & 34.4 & 23.8 & 30.8 & 22.2 & 26.4 & 29.1
  & 35.0 & 31.4 \\
VG-LLM
  & 55.2 & 32.3 & 58.5 & 57.1 & 51.6 & 32.2
  & 56.0 & 36.3 & 32.0 & 30.3 & 32.1 & 29.1
  & 27.9 & 38.4 \\
DSR Suite-Model
  & 87.0 & 73.8 & 84.1 & 73.8 & 72.0 & 35.5
  & 75.8 & 76.1 & 77.7 & 60.6 & 37.1 & 35.1
  & 46.4 & 58.9 \\

GeoSR
    & 88.0 & 84.4 & 87.8 & 83.3 & 78.5 & 41.0
    & 79.5 & 83.0 & 83.8 & 76.5 & 48.9 & 39.8
    & 52.9 & 66.1\\

4DThinker
  & 84.3 & 79.7 & 82.3 & 84.0 & 67.0 & 47.8
  & 67.0 & 82.8 & 79.5 & 62.9 & 44.2 & 41.0
  & 48.5 & 62.0 \\

\addlinespace[4pt]
\midrule
\multicolumn{15}{c}{\textbf{\textit{Ours (Based on Qwen3-VL-4B)}}} \\[1pt]
Qwen3-VL-4B
    & 34.1 & 31.7 & 29.5 & 37.5 & 14.3 & 30.4
    & 22.5 & 29.5 & 35.4 & 20.8 & 19.6 & 20.6
    & 35.7 & 27.7 \\
  \quad \textit{+SFT}
    & 89.0 & 71.4 & 82.1 & 72.5 & 62.6 & 43.5
    & 77.5 & 76.1 & 79.7 & 49.0 & 38.4 & 38.8
    & 46.0 & 58.8 \\

  \quad \textit{+SFT+Raw}
    & 89.0 & 73.0 & 85.9 & 68.8 & 65.9 & 46.1
    & 77.5 & 78.4 & 82.3 & 59.4 & 41.3 & 33.3
    & 48.1 & 60.2 \\

  \quad \textit{+SFT+Raw+Delta}
    & 91.5 & 76.2 & 91.0 & 66.2 & 74.7 & 42.6
    & 77.5 & 81.8 & 81.0 & 75.0 & 50.7 & 37.6
    & 43.0 & 62.7 \\

\midrule
\multicolumn{15}{c}{\textbf{\textit{Ours (Based on Qwen3-VL-8B)}}} \\[1pt]

Qwen3-VL-8B
    & 23.5 & 24.6 & 42.6 & 29.7 & 27.9 & 33.8
    & 18.1 & 28.4 & 34.5 & 24.2 & 22.1 & 27.9
    & 33.5 & 28.7 \\
  \quad \textit{+SFT}
    & 89.0 & 82.5 & 88.5 & 78.8 & 70.3 & 47.8
    & 74.8 & 83.0 & 86.1 & 63.5 & 42.0 & 36.4
    & 43.4 & 62.0 \\
  \quad \textit{+SFT+Raw}
   & 89.0 & 87.3 & 89.7 & 77.5 & 80.2 & 43.5
   & 76.6 & 88.6 & 86.1 & 83.3 & 58.7 & 38.2
   & 41.3 & 65.8 \\
  \quad \textit{+SFT+Raw+Delta}
   & 89.0 & 82.5 & 92.3 & 77.5 & 80.2 & 44.3
   & 76.6 & 90.9 & 83.5 & 84.3 & 53.6 & 38.2
   & 52.0 & \textbf{67.2} \\

\bottomrule
\end{tabular}
}
\vspace{-1.2em}
\end{table*}

%% file: Sec/exp.tex
\section{Experiments}
\label{sec:experiments}

\subsection{Experimental setup}

\paragraph{Benchmark and metrics.}
We evaluate on DSR-Bench~\citep{zhou2026learning}, a multiple-choice benchmark
for fine-grained dynamic spatial reasoning from monocular videos.  It contains
13 subtasks: six spatial attributes---distance, direction, orientation, speed,
speed comparison, and direction prediction---are evaluated under absolute and
relative viewpoints, together with a non-template subset.  We report accuracy
for each subtask and overall sample-level accuracy.

\paragraph{Implementation details.}
Our default model uses Qwen3-VL-8B-Instruct~\citep{Qwen3-VL}
as the VLM and frozen Wan2.1-T2V-1.3B~\citep{wan2025wan}
as the VGM. We extract VGM activations offline from zero-indexed
block 20 under an empty-text condition and diffusion timestep of
300.  We train on the DSR-train split for one epoch
with a global batch size of 32 and a learning rate of $2\times 10^{-7}$.  The VGM and native VLM
visual encoder remain frozen. We use AdamW with a cosine
learning-rate schedule, a warmup ratio of 0.03, and weight decay of 0.01.
All experiments are conducted on $8\times$A100 GPUs.

\subsection{Results and Analysis}
\label{sec:sota_comparison}

Table~\ref{tab:dsrbench} evaluates whether VGM representations improve over
strong public and proprietary systems.  \method with Qwen3-VL-8B reaches
\textbf{67.2\%}, surpassing the previous best GeoSR~\citep{zhang2026make},
which uses the 3D foundation model $\pi^3$~\citep{wang2026pi},
establishing a new state of the art on DSR-Bench.  The
gain is not explained by supervised fine-tuning alone: Qwen3-VL-8B SFT obtains
62.0\%, raw transfer raises accuracy to 65.8\%, and the full raw--delta method
reaches 67.2\%.  The 4B backbone shows the same ordering, with raw--delta fusion
reaching 62.7\% compared with 58.8\% for SFT.  These cross-scale results show
that the transferred representation improves both backbones under the
protocol.

\subsection{Ablation studies}

\paragraph{Raw and delta branches.}
Table~\ref{tab:component_ablation} evaluates the two transferred signals.
Raw and delta features independently improve
over SFT, while their combination performs best, supporting their
complementarity.

\paragraph{VGM feature block.}
Table~\ref{tab:vgm_block} varies the extraction layer while keeping the
raw--delta pathway, diffusion timestep, and training protocol fixed.  Block 20
performs best, while both shallower and deeper
features are weaker.  This indicates that transferable evidence is concentrated
in an intermediate-to-late representation rather than increasing monotonically
with network depth.

\paragraph{Diffusion timestep.}
Table~\ref{tab:diffusion_timestep} examines how the diffusion timestep shapes
the transferred representation. Performance is lower for the clean and lightly
noised states, peaks at an intermediate timestep, and then gradually declines
as the corruption becomes stronger. This non-monotonic trend suggests that an intermediate
noising state may better balance the preservation of input evidence with the
VGM's learned denoising prior.

\begin{table*}[t]
  \centering
  \input{Tabs/tab_component_ablation}\hfill
  \input{Tabs/tab_vgm_block}\hfill
  \input{Tabs/tab_diffusion_timestep}
  \vspace{-0.6em}
\end{table*}

%% file: Tabs/tab_component_ablation.tex
\begin{minipage}[t]{0.39\textwidth}
  \vspace{0pt}
  \centering
  \small
  \setlength{\tabcolsep}{4.5pt}
  \caption{\textbf{Raw--delta components.}}
  \label{tab:component_ablation}
  \begin{tabular}{lccc}
    \toprule
    \textbf{Variant} & \textbf{Raw} & \textbf{Delta}
       & \textbf{Acc} \\
    \midrule
    Base & -- & -- & 28.70 \\
    SFT & -- & --  & 62.00 \\
    Raw only        & \checkmark & -- & 65.80  \\
    Delta only      & -- & \checkmark  & 64.32 \\
    \method         & \checkmark & \checkmark
      & \textbf{67.20}  \\
    \bottomrule
  \end{tabular}
\end{minipage}

%% file: Tabs/tab_vgm_block.tex
\begin{minipage}[t]{0.25\textwidth}
  \vspace{0pt}
  \centering
  \small
  \setlength{\tabcolsep}{7pt}
  \caption{\textbf{VGM block.}}
  \label{tab:vgm_block}
  \begin{tabular}{lc}
    \toprule
    \textbf{Block} & \textbf{Acc} \\
    \midrule
     5 & 65.46 \\
    10 & 65.39 \\
    15 & 66.64 \\
    \textbf{20} & \textbf{67.20} \\
    29 & 66.43 \\
    \bottomrule
  \end{tabular}
\end{minipage}

%% file: Tabs/tab_diffusion_timestep.tex
\begin{minipage}[t]{0.30\textwidth}
  \vspace{0pt}
  \centering
  \small
  \setlength{\tabcolsep}{5pt}
  \caption{\textbf{VGM timestep.}}
  \label{tab:diffusion_timestep}
  \begin{tabular}{lc}
    \toprule
    \textbf{Timestep} & \textbf{Acc} \\
    \midrule
    No noise & 65.60 \\
    $t=100$ & 65.04 \\
    $t=300$
      & \textbf{67.20} \\
    $t=600$ & 66.92 \\
    $t=900$ & 66.43 \\
    \bottomrule
  \end{tabular}
\end{minipage}

%% file: Sec/conclusion.tex
\section{Conclusion}
\label{sec:conclusion}

We introduce \method, which extracts intermediate
representations from a frozen video generation model and transfers
them into a VLM through complementary raw and delta
branches. Experiments on DSR-Bench show that generative representations consistently improve
VLM performance. These results suggest that video generation models can serve
not only as synthesis models, but also as a
valuable source of representations for reasoning about the evolving physical
world.